\documentclass[graybox]{svmult}

\usepackage{mathptmx}       
\usepackage{helvet}         
\usepackage{courier}        
\usepackage{type1cm}        
\usepackage{makeidx}         
\usepackage{graphicx}        
\usepackage{multicol}        
\usepackage[bottom]{footmisc}

\usepackage{siunitx}
\usepackage{amsfonts}
\usepackage{subcaption}
\usepackage{booktabs}

\makeindex             

\begin{document}

\title*{Mixed Cooperative-Competitive Communication Using Multi-Agent Reinforcement Learning}
\titlerunning{Mixed Cooperative-Competitive Communication Using MARL}

\author{Astrid Vanneste, Wesley Van Wijnsberghe, Simon Vanneste, Kevin Mets, Siegfried Mercelis, Steven Latré, Peter Hellinckx}
\authorrunning{Astrid Vanneste et al.}

\institute{
    Astrid Vanneste\textsuperscript{\rm 1},
    Wesley Van Wijnsberghe\textsuperscript{\rm 1},
    Simon Vanneste\textsuperscript{\rm 1},
    Kevin Mets\textsuperscript{\rm 2},
    Siegfried Mercelis\textsuperscript{\rm 1},
    Steven Latré\textsuperscript{\rm 2},
    Peter Hellinckx\textsuperscript{\rm 1}
    \at{
    \textsuperscript{\rm 1}
    University of Antwerp - imec
    IDLab - Faculty of Applied Engineering
    Sint-Pietersvliet 7, 2000 Antwerp, Belgium
    \\\textsuperscript{\rm 2}
    University of Antwerp - imec
    IDLab - Department of Computer Science
    Sint-Pietersvliet 7, 2000 Antwerp, Belgium
    } \\
    \email{
        astrid.vanneste@uantwerpen.be,
        wesley.vanwijnsberghe@student.uantwerpen.be,
        simon.vanneste@uantwerpen.be,
        kevin.mets@uantwerpen.be,
        siegfried.mercelis@uantwerpen.be,
        steven.latre@uantwerpen.be,
        peter.hellinckx@uantwerpen.be
}}
%
%

\thispagestyle{empty} 
\section*{Copyright Notice}
This is a preprint of the following chapter: Astrid Vanneste, Wesley Van Wijnsberghe, Simon Vanneste, Kevin Mets, Siegfried Mercelis, Steven Latr\'e, Peter Hellinckx, Mixed Cooperative-Competitive Communication Using Multi-agent Reinforcement Learning, published in Advances on P2P, Parallel, Grid, Cloud and Internet Computing. 3PGCIC 2021. Lecture Notes in Networks and Systems, vol 343., edited by Leonard Barolli, 2021, Springer reproduced with permission of Springer Nature Switzerland AG. The final authenticated version is available online at: http://dx.doi.org/10.1007/978-3-030-89899-1\_20.

{
	\let\clearpage\relax
	\maketitle
}

\abstract*{By using communication between multiple agents in multi-agent environments, one can reduce the effects of partial observability by combining one agent's observation with that of others in the same dynamic environment. While a lot of successful research has been done towards communication learning in cooperative settings, communication learning in mixed cooperative-competitive settings is also important and brings its own complexities such as the opposing team overhearing the communication. In this paper, we apply differentiable inter-agent learning (DIAL), designed for cooperative settings, to a mixed cooperative-competitive setting. We look at the difference in performance between communication that is private for a team and communication that can be overheard by the other team.
Our research shows that communicating agents are able to achieve similar performance to fully observable agents after a given training period in our chosen environment. Overall, we find that sharing communication across teams results in decreased performance for the communicating team in comparison to results achieved with private communication.
}

\abstract{By using communication between multiple agents in multi-agent environments, one can reduce the effects of partial observability by combining one agent's observation with that of others in the same dynamic environment. While a lot of successful research has been done towards communication learning in cooperative settings, communication learning in mixed cooperative-competitive settings is also important and brings its own complexities such as the opposing team overhearing the communication. In this paper, we apply differentiable inter-agent learning (DIAL), designed for cooperative settings, to a mixed cooperative-competitive setting. We look at the difference in performance between communication that is private for a team and communication that can be overheard by the other team.
Our research shows that communicating agents are able to achieve similar performance to fully observable agents after a given training period in our chosen environment. Overall, we find that sharing communication across teams results in decreased performance for the communicating team in comparison to results achieved with private communication.
}


\section{Introduction}
\label{sec:introduction}
Research has demonstrated that single-agent reinforcement learning (RL) can successfully be applied to tackle challenging problems in a variety of applications \cite{Mnih,Silver}. However, several applications (e.g. multiplayer games) require interaction between multiple agents in order to reach a certain goal. The tasks that we consider in our application environment are mixed cooperative-competitive, partially observable multi-agent decision making problems in which the agents of a team share a common goal. These goals are competitive across teams. Due to the joint observability of the underlying Markov state, each individual agent only knows its own observation and consequently lacks knowledge of the full state of the dynam+ic environment. With a view to optimise the decision-making process of each individual member of a team, we need a technique for these agents to share their state information with each other in order to achieve a greater knowledge about the full underlying state.
Recently, interest has been directed to a research domain within multi-agent reinforcement learning (MARL) that opens the door to a new perspective on coordinated behaviour of multiple RL agents in a dynamic environment \cite{Foerster2016, Sukhbaatar2016}. This domain focuses its research on agents that use machine learning principles to learn to communicate amongst themselves while simultaneously learning the behaviour needed to achieve a pre-defined common goal. In real-world environments, individual agents generally do not have full observability of the surrounding environment. Therefore, it is essential that these agents learn a protocol to share their individual observations, so that each agent can generate a more complete representation of the environment. The more information an agent has about its immediate environment, the better it is able to make good decisions in order to achieve the common goal.
The aim of this paper is to investigate current state-of-the-art communication learning methods in a mixed cooperative-competitive setting. We want to look into the effects of sharing these communication messages with the opposing team. We continue this paper by discussing the state of the art on multi-agent communication in section \ref{sec:related_work}. Section \ref{sec:background} contains some background information regarding RL. In section \ref{sec:methods}, we describe the used methods. Our experiments are presented in section \ref{sec:experiments}. We discuss the results of our research in section \ref{sec:results}.
To finalise this paper, we discuss our conclusions and some future work regarding multi-agent communication in section \ref{sec:conclusion}.


\section{Related Work}
\label{sec:related_work}
In this section, we focus on the related work concerned with mixed cooperative-competitive MARL and the learning of communication protocols in MARL.

Research within MARL is mostly focused on cooperative tasks. However, Lowe et al. \cite{Lowe2017} presented a set of tasks which are fully cooperative or mixed cooperative-competitive. They present their method, multi-agent deep deterministic policy gradient (MADDPG). This method is able to tackle these tasks, including some communication learning tasks.

Foerster et al. \cite{Foerster2016} and Sukhbaatar et al. \cite{Sukhbaatar2016} presented the foundations in the field of communication learning. Sukhbaatar et al. \cite{Sukhbaatar2016} explores a neural model, called CommNet, that uses continuous communication for fully cooperative tasks. In this model, communication between agents is achieved by sharing the agents' internal hidden states with the other agents. Agents learn a communication protocol by allowing gradients to flow through the communication channel. This results in improved performance over non-communicative agents.
Two different approaches for learning discrete communication are proposed by Foerster et al. \cite{Foerster2016}: Reinforced Inter-Agent Learning (RIAL) and Differentiable Inter-Agent Learning (DIAL). RIAL uses deep Q-learning \cite{Mnih} for both the action and the communication policy. There is no direct feedback from other agents on the communication of a certain agent, which makes it difficult for the agents to agree on a communication policy. DIAL addresses this limitation by allowing gradients to flow between agents through the communication channel. The gradients give richer feedback, reduce the required amount of learning by trial and error, and ease the discovery of effective protocols. They are able to learn discrete communication protocols by applying a discretize regularize unit (DRU) on the messages.
Following these fundamental works of Sukhbaatar et al. and Foerster et al. more research in the field of communication learning emerged. BiCNet\cite{peng2017bicnet} and ACCNet \cite{mao2017accnet} are both actor-critic approaches to communication learning. Vanneste et al. \cite{vanneste2020value} combined the work of Foerster et al. \cite{Foerster2016} and Sunehag et al. \cite{sunehag2018value} to improve the performance of DIAL. MACC \cite{vanneste2020learning} is an approach with a centralized critic that applies counterfactual reasoning to learn to communicate. Recently, there has also been more attention to directional communication \cite{das2018tarmac}\cite{ding2021learning}\cite{jiang2018attentional} .


\section{Background}
\label{sec:background}
In this section we discuss some required background information on RL.
In single-agent RL, the agents' environment can be described by a Markov decision process (MDP)\cite{howard1960dynamic}. A MDP is a tuple ($S$, $U$, $f$, $\rho$) where $S$ is the set of environment states $s$, $U$ is the set of agent actions $u$, $f : S \times U \times S \rightarrow [0,1]$ is the state transition probability function, and $\rho : S \times U \rightarrow \mathbb{R}$ is the reward function.
As a result of the action $u_{t} \in U$, the environment changes its state from $s_{t}$ to some $s_{t+1} \in S$ according to the state transition probabilities defined by $f$. After this transition, the agent receives a scalar reward $r_{t+1}$ $\in$ $\mathbb{R}$ according to the reward function $\rho$. A policy $\pi$ describes the behaviour of the agent, and therefore specifies how the agent chooses its actions given the observed state.
The task of the agent is to learn a policy to maximize its long-term expected reward.
When agents do not receive the full state of the environment but only a limited state observation $o$, the multi-agent system is described using a partially observable MDP (POMDP).

MARL problems where all agents receive the full state are modelled using multi-agent MDPs (MMDP). In contrast, problems where agents only receive a limited state observation are described by Oliehoek et al. \cite{dec-POMDP} as decentralized POMDPs (dec-POMDP). Decentralized MDPs (dec-MDP) are a subset of dec-POMDPs for jointly observable problems. In these jointly observable problems, the full state $s_t$ can be composed by combining all individual observations of the agents \cite{goldman2003optimizing} \cite{dec-POMDP}. In our experiments we try to tackle these dec-MDPs, since the agents in our multi-agent mixed cooperative-competitive setup lack the necessary information to achieve their goal, but are able to generate full observability by exchanging state information.


\section{Methods}
\label{sec:methods}
In this section, we present our approach for learning communication protocols in a mixed cooperative-competitive environment where two teams compete against each other. In our experiments we will be using a predator-prey environment where the predators need to capture a specific prey. Only the other predator knows which prey needs to be targeted. This results in a need for communication. The environment is explained in more detail in Section \ref{sec:environment}. The agents of the two teams are different. The team of predators will be able to communicate while the team of prey are not.

\subsection{Prey Agents}

The team of prey agents is not allowed to communicate. We opted to train our prey agents with independent Q-learing (IQL)\cite{Tan1993MultiAgentRL}. All prey agents share the same parameters, which are synchronized after each epoch, to achieve better results. The prey agents will not be able to gain full observability if they cannot overhear the communication of the predators since they do not know which of the predators is targeting them.

\subsection{Predator Agents}

\label{sec:DIAL-SCN}
\begin{figure}[h]
    \centering
    \includegraphics[width = \linewidth]{{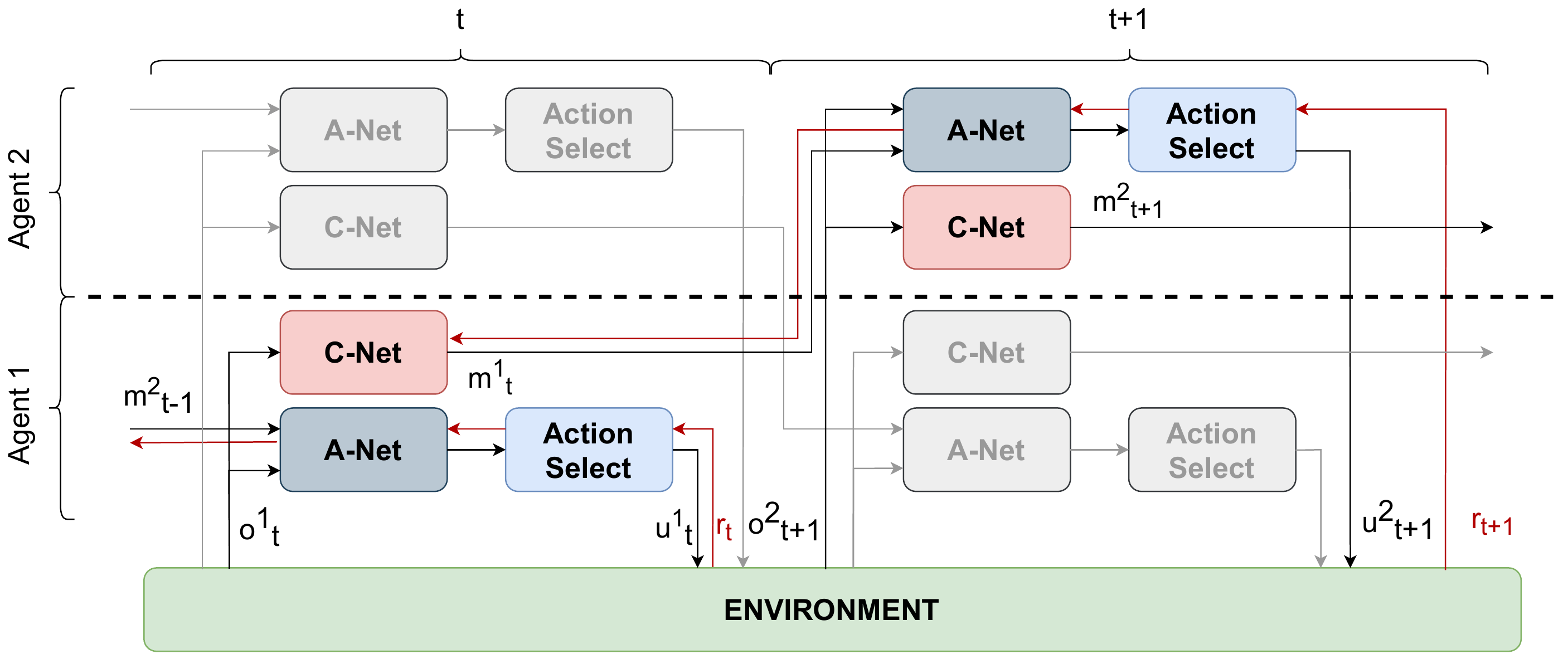}}
    \caption{DIAL without DRU and with a separate action and communication policy}
    \label{fig:comm_network}
\end{figure}

The team of predator agents will be able to communicate with each other. The method we use is inspired by DIAL as described by Foerster et al. \cite{Foerster2016} and can be seen in Fig. \ref{fig:comm_network}. The key insight behind DIAL is the possibility of pushing gradients from one agent to another through the communication channel. The gradient flow from one agent to the other acts as a form of feedback, which does not only reduce the amount of learning by trial and error but also eases the discovery of effective communication protocols.
Comparing our method to the original method presented by Foerster et al., there are two main difference. First, we have split the neural network of the action and the communication policy. This allows us to use a very simple network for the communication policy, making it easier to learn. Secondly, we chose not to include the DRU, since we observed better results when using a ReLU activation instead. This causes communication to be continuous at both execution and training time.
The communication network, which is called the C-Net, is a neural network which outputs a vector message based on the environment observation it receives as input. The vector message is passed to the other agent.
The action network, called the A-Net, calculates the Q-values based on the received vector message from its teammate and the observation received from the environment. These Q-values are passed on to the action select, which selects the optimal action based on these Q-values. The predator agents share their parameters to improve performance. The parameters are synchronized after each epoch.


\section{Experiments}
\label{sec:experiments}
In order to evaluate our approach, we perform a set of experiments in a mixed cooperative-competitive environment. In this section we discuss the specification of the constructed environment and our agent setup.

\subsection{Environment}
\label{sec:environment}

In our research we use a combination of the "simple tag" and "simple reference" environments provided by OpenAi \cite{Lowe2017, Mordatch} to represent our mixed cooperative-competitive multi-agent environment. The environment is a predator-prey environment where the predators have to catch a specific prey and the prey have to avoid the predator that is targeting them. In our experimental setup, only the two predator agents will be able to communicate with each other. This communication is a key factor in their performance, since every predator agent is assigned a specific enemy target, which is only known by its teammate. In order for these agents to reach the common goal of getting as close as possible to their respective target prey, they have to learn a way to communicate the targets to each other. The goal of the prey agents is to maximise their combined distance from the predators assigned to catch them. Since there are two possible targets for the predator agents, the size of the communication space is one.

\subsection{Model Architecture}

The predators use the C-Net to learn the message policy. The C-Net is a fully connected neural network with no hidden layers. In our setup we do not need hidden layers, because our agents have to directly communicate certain information from their observation. It accepts an observation input and outputs the vector message. This network is effectively a linear mapping between the input and output.

The A-net accepts an observation vector input from the environment and a message vector input from the agent's teammate. The inputs are concatenated and passed through two fully connected hidden layers with a size of 256 and 512 respectively. This network is the only network used by the prey agents because they are not required to learn communication.

We perform our experiments in the RLlib framework \cite{liang2018rllib}. In our experiments we perform 30 steps in a single episode. The learning rate is set at 0.0005, while we choose a discount factor of 0.97. The networks are updated every epoch, consisting of 50 episodes. We use a training batch size of 200.

\subsection{Experimental Setup}
In order to evaluate the effectiveness of the learned communication, we choose to train our agents in four distinct configurations.

\begin{itemize}
    \item \textbf{No Communication}: The predator agents are not allowed to learn a communication protocol. They have to base their actions solely on the environmental observation, which lacks target information for a single predator agent. Here, both teams will be trained using IQL\cite{Tan1993MultiAgentRL} with a partially observable environment.
    \item \textbf{Full Observability}: Instead of observing the teammate target, predators observe their own target. They have all the necessary information required to maximize the team reward without the need to communicate with its teammate. Since there is no need for communication, both teams are trained using IQL \cite{Tan1993MultiAgentRL}.
    \item \textbf{Private Communication}: The predator agents communicate with each other, while the prey agents are not able to observe this communication.
    \item \textbf{Public Communication}: The predator agents communicate with each other, while the prey agents can hear this communication. They effectively observe the communication of the predators and if they are able to use the predator communication to learn which prey agent(s) is/are targeted during each episode, they can use this extra information to increase their performance. This is the only setting where the prey are able to determine which prey is being targeted by which predator.
\end{itemize}


\section{Results}
\label{sec:results}
\begin{figure}[]
    \centering
    \includegraphics[width=\textwidth]{{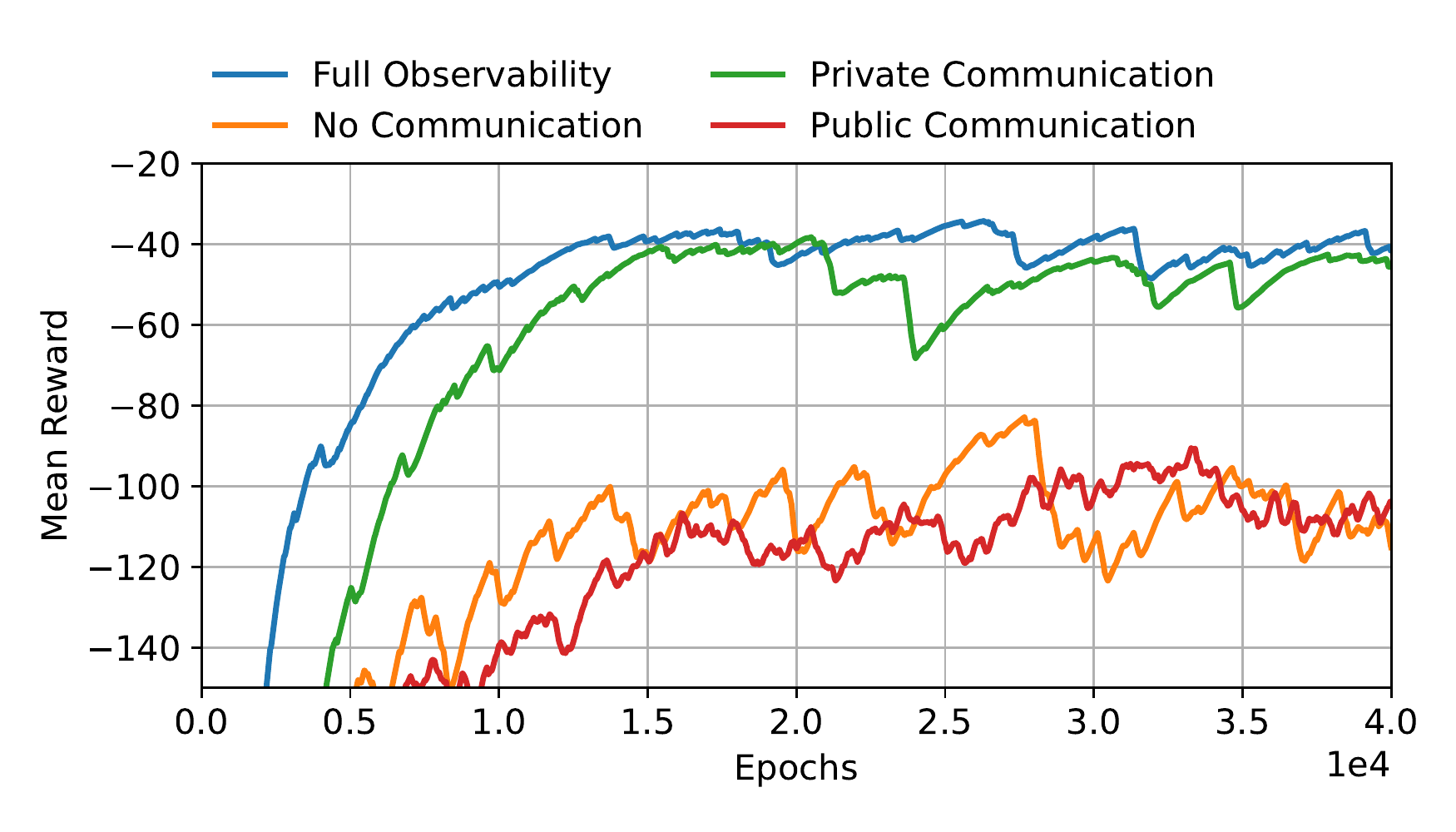}}
    \caption{Mean reward of the predators for each of the experiment settings smoothed with an exponentially weighted moving average filter with an alpha value of 0.0005.}
    \label{fig:predator_rewards}
\end{figure}

In this section, we evaluate the results of the four proposed experiments from the previous section. Because of the fact that only a single team, namely the team of predators, is able to learn communication, we choose to concentrate on their results. In each of our experimental setups the prey merely act as an opponent in order to fulfill the competitive aspect of our research. Across most experiments, they share the same configuration apart from the prey in the public communication experiment, who differ by the ability to observe the predator communication. Due to the competitive setup of our environment and the way the rewards are calculated, the performance of the prey is inversely proportional to the performance of the predators.
In Fig. \ref{fig:predator_rewards} the results from five training runs for the mean predator rewards are shown. It shows the average reward received during an episode per epoch for the entire training run. We choose to smooth the results by applying an exponentially weighted moving average filter with an alpha value of 0.0005 in order to ease the comparison of the different experiments. Table \ref{table:std} shows the average reward and the average standard deviation across the five training runs. We see that overall the full observability setting performs best with the private communication settings achieving very similar results. They both also show the lowest standard deviation. The public communication setting however shows a much worse performance, close to the performance of the no communication setting. However, the public communication setting is able to perform much better than the no communication setting when we look at the average peak performance accross the different training run of each of the settings. The peak performance shows the maximum reward that is achieved in a training run. This can be seen in Table \ref{table:peak_perf}. We see that the private communication setting performs best closely followed by the full observability setting. However, the public communication setting performs a lot better than the no communication setting. This is to be expected since the predators are able to communicate about their target, but they cannot perform as well as in the private communication setting because the prey can overhear the communication and determine which of the predators is targeting them.

\begin{table}[]
\centering
\begin{tabular}{c|c|c}
                        & Average Reward    & Standard Deviation \\ \hline
Full Observability      & -50.93            & 27.65              \\
No Communication        & -119.81           & 69.81              \\
Private Communication   & -64.49            & 42.05              \\
Public Communication    & -129.20           & 75.69              \\
\end{tabular}
\caption{Overview of the average reward and the average standard deviation across the five training runs for each of the experiment settings}
\label{table:std}
\end{table}

\begin{table}[]
\centering
\begin{tabular}{c|c|c}

                            & Average Peak Performance & Standard Deviation of the Peak Performance \\ \hline
Full Observability          & -18.6                    & 1.02                                       \\
No Communication            & -43.6                    & 1.96                                       \\
Private Communication       & -17.6                    & 0.49                                       \\
Public Communication        & -24.0                    & 1.26                                       \\
\end{tabular}
\caption{Overview of the average peak performance and the standard deviation on the average peak performance across the five training runs for each of the experiment settings}
\label{table:peak_perf}
\end{table}

\begin{figure}
    \begin{subfigure}{0.49\textwidth}
        \includegraphics[width=\textwidth]{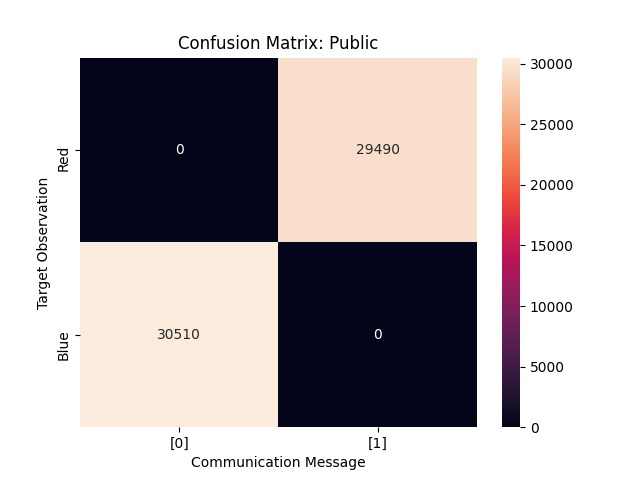}
        \caption{Public communication}
        \label{fig:cm_public}
    \end{subfigure}
        \hspace*{\fill}
    \begin{subfigure}{0.49\textwidth}
        \includegraphics[width = \linewidth]{{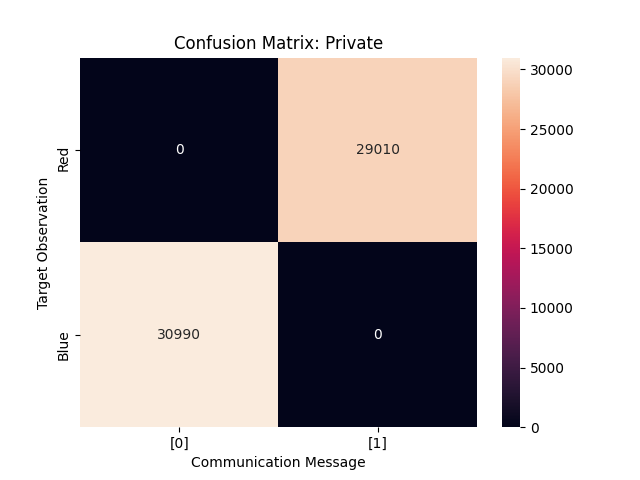}}
        \caption{Private communication}
        \label{fig:cm_private}
    \end{subfigure}
    \caption{Confusion matrix from the communication policy of the predator agents}
    \label{fig:confusion_matrix}
\end{figure}

Since there is a significant difference in performance between the private and public communication settings, it is interesting to look at their communication policies. In Fig. \ref{fig:confusion_matrix}, the communication protocols are presented as a confusion matrix. In this confusion matrix, we have discretised the communication by applying a threshold at zero for the messages. Here, we can see which message is sent in function of which prey the other predator has to target. We can clearly see that in both settings the agents are able to learn a valid communication protocol. Therefore, we can conclude that the difference in performance is not caused by an invalid communication protocol in the public communication setting but by the fact that the prey can overhear communication.

\section{Conclusion and Future Work}
\label{sec:conclusion}
In this paper, we investigated the effect of learning to communicate in mixed cooperative-competitive environments where it is possible for communication to be heard by the opposing team. In order to perform the required research, we use a technique for of the same team to learn a communication protocol inspired by the state of the art as presented by Foerster et al. \cite{Foerster2016}. This technique was tested in a mixed cooperative-competitive environment that acted as a predator-prey environment where the agents are rewarded based on the distance to their respective target. We compare the achieved results using this proposed technique for four distinct configurations. We can conclude that the predators in the private experiment are able to achieve a similar performance to predators with full observability of the environment. When looking at the predators at peak performance, we acquired the knowledge that the predators in the public setting performed worse than in the private setting. Our experiments showed that using DIAL \cite{Foerster2016}, predator agents are able to learn a valid communication protocol for both public and private configurations. This leads to the conclusion that in scenarios where communication is shared across teams, performance will decline significantly. In the future, it would be interesting to look at a scenario with more agents, since we limited the size of the teams to two agents. In our results, we saw that public communication makes it easier for the opponents to anticipate on the actions of the agents. To prohibit this, it would be interesting to look at techniques that will allow the agents to encrypt their messages and to reason about the behaviour of opposing agents as a reaction to the communication.

\begin{acknowledgement}
This work was supported by the Research Foundation Flanders (FWO) under Grant Number 1S12121N and Grant Number 1S94120N.
We gratefully acknowledge the support of NVIDIA Corporation with the donation of the Titan Xp GPU used for this research.
\end{acknowledgement}

\bibliographystyle{spmpsci}
\bibliography{ref}

\end{document}